\documentclass{elsarticle}
\usepackage{lineno}
\usepackage{amssymb}
\usepackage{latexsym}
\usepackage{amsfonts}
\usepackage{amsmath}
\usepackage{algorithm}
\usepackage{algpseudocode}
\usepackage{subfigure} 

\modulolinenumbers[5]

\journal{}

\begin{document}

\begin{frontmatter}

\title{Minimum description length as an objective function for non-negative matrix factorization}

\author{Steven Squires, Adam Pr\"{u}gel Bennett and Mahesan Niranjan}
\address{University of Southampton, University Road, Southampton, UK}

\begin{abstract}
Non-negative matrix factorization (NMF) is a dimensionality reduction technique which tends to produce a sparse representation of data. Commonly, the error between the actual and recreated matrices is used as an objective function, but this method may not produce the type of representation we desire as it allows for the complexity of the model to grow, constrained only by the size of the subspace and the non-negativity requirement. If additional constraints, such as sparsity, are imposed the question of parameter selection becomes critical. Instead of adding sparsity constraints in an ad-hoc manner we propose a novel objective function created by using the principle of minimum description length (MDL). Our formulation, MDL-NMF, automatically trades off between the complexity and accuracy of the model using a principled approach with little parameter selection or the need for domain expertise. We demonstrate our model works effectively on three heterogeneous data-sets and on a range of semi-synthetic data showing the broad applicability of our method.
\end{abstract}


\end{frontmatter}



\section{Introduction}

Non-negative matrix factorisation (NMF) is a popular linear dimensionality reduction technique. Its ability to produce a sparse and parts based representation, in contrast to principal component analysis (PCA) which is variance preserving, has been exploited in a range of applications~\cite{devarajan2006nonnegative,lee1999learning,wang2011community}. The aim of NMF is to take a non-negative data matrix $\textbf{V} \in \mathbb{R} ^{m \times n}$ with $m$ dimensions and $n$ data points and find two non-negative matrices: $\textbf{W} \in \mathbb{R} ^{m \times r}$ and $\textbf{H} \in \mathbb{R} ^{r \times n}$, such that $\textbf{V}\approx \textbf{WH}$ and (usually) $r\ll m$ and $r\ll n $. The columns of $\textbf{W}$ represent the axes of the new subspace and each column of $\textbf{H}$ holds the coefficients of each data-point in the new subspace. The standard method for finding the matrices $\textbf{W}$ and $\textbf{H}$ is to choose an objective function and then attempt to minimise it. One of the most common choices is the Frobenius norm: $f=||\textbf{V}-\textbf{WH}||^2_{\text{Fro}}$, but other options such as measures similar to the Kullback-Leibler divergence have also been used~\cite{lee2001algorithms}.

One of the useful features of NMF is that it can produce a sparse representation of the data which, amongst other benefits, makes it easier to interpret. This sparsity can manifest in either, or both, of the factorized matrices. Sparsity is imposed by NMF because of the non-negativity constraint which means all the latent structure is positive and so is the summation of that structure to recreate each data-point. The columns of the $\textbf{W}$ matrix can often be interpreted as parts of the data, especially as the features occupy the same positive space as the input matrix. Again, this is in contrast to PCA where the subspace the data is projected onto tends not to be purely positive and therefore is difficult to interpret as features of a non-negative input matrix. Neither does PCA normally produce sparse solutions, instead producing holistic representations which preserve variance.

The seminal paper by Lee and Seung~\cite{lee1999learning}, which provided the surge of interest in NMF (although other authors had conducted significant work in the field previously~\cite{paatero1994positive}), relied on the constraints from the small subspace size and non-negativity to provide the sparse solutions. However, other authors, in particular Hoyer~\cite{hoyer2004non}, showed that for some problems NMF does not produce sparse solutions and it can be useful to impose sparsity explicitly.

Hoyer~\cite{hoyer2004non} introduced additional sparsity parameters through a projection operator which constrains the $\textbf{W}$ and $\textbf{H}$ matrices to have a defined level of sparseness. The formulation of Hoyer works well if there is a specific goal in mind. If you want to extract understandable features you might want to impose significant levels of sparsity on $\textbf{W}$ while if you are more interested in a representation closer to clustering, imposing sparsity constraints on $\textbf{H}$ can be effective. These choices are domain dependent and require ad-hoc choices. However, if there is no prior knowledge or domain expertise, how do you decide the level of sparsity to impose on the factorized matrices and the distribution of the imposed sparsity between the two matrices? At a more fundamental level we ask whether the minimisation of the error alone is an optimal strategy.

With no additional constraints NMF algorithms attempt to minimise the error between the recreated and original matrices, or for other objective functions, such as the Kullbeck-Leibler divergence, to maximise the similarity in distribution between the actual and recreated matrices. In many situations we would expect this to result in over-fitting of the data - modelling noise rather than signal, and resulting in a model which would generalise poorly. The algorithm aims to maximise the similarity between the actual and recreated matrices with no regard for how complex the model being created is. As this is unsupervised learning the algorithm sees all the data and cannot be easily tested on unseen data. 

There are two features of the formulation that contend with overfitting: 1) the small size of the subspace $r$, which means that it is likely real latent structure will be uncovered; 2) the non-negativity constraint which works to prevent overfitting by preventing the subtraction of structure, so providing an emphasis on sparse models which do not include unwanted structure. The choice of $r$ has been investigated in the literature~\cite{squires2017rank} and its importance is clear when considering that if $r=m$ or $r=n$ then the model can perfectly recreate the actual data. The quality of the fit increases continuously with an increasing $r$. The non-negativity constraint can produce sparse solutions which act against overfitting but, without additional constaints, may not be sufficient as demonstrated by Hoyer~\cite{hoyer2004non}. The objective function without sparsity constraints may be biased in favour of improving accuracy with less consideration of the level of complexity of the model (which causes the overfitting). The imposition of sparsity constraints is an effective but ad-hoc and arbitrary method to impose an additional constraint on the model becoming too complex. A more fundamental approach is desirable.

A Bayesian approach to NMF has been demonstrated by, for example~\cite{schmidt2009bayesian}, which overcomes some of these related issues, especially the ad-hoc nature of additional sparsity constraints. Other groups have also applied a Bayesian formulation~\cite{blei2010bayesian}, including to the choice of $r$. Bayesian approaches have great value in NMF but they still require knowledge of the domain to enable prior distributions to be chosen. Other related work has been performed using particle filtering ideas~\cite{wang2013online}.

We propose an alternative objective function using a formulation based on minimum description length (MDL). This formulation allows an automatic and natural trade-off between accuracy and complexity of the model without the need for considerable domain expertise. Previous work has used MDL ideas to estimate good choices of $r$~\cite{squires2017rank} in NMF but, to our knowledge, no previous work has used MDL principles to guide the formation of the $\textbf{W}$ and $\textbf{H}$ matrices.

The remainder of this paper is organised as follows. In Section~\ref{sec:mdl} we present the principles behind our application of MDL to the NMF objective function; in Section~\ref{sec:methods} we explain our methodology and algorithm; in Section~\ref{sec:results} we demonstrate the empirical effectiveness of our model on real and semi-synthetic data and finally in Section~\ref{sec:conclusions} we draw conclusions and propose future directions of research.


\section{Minimum description length}
\label{sec:mdl}
MDL is a method which produces an automatic trade-off between the complexity and accuracy of a model~\cite{rissanen1978modeling,wallace1968information,barron1998minimum}. Considering a message sent to a receiver, the MDL principle is that the encoding of the message that can be described by the shortest code is the best choice, as long as it enables the message to be recreated with perfect (to a pre-agreed precision) accuracy by the receiver. This is essentially equivalent to the best compression of the data, as a maximally compressed data-set will also be cheaper to transmit as a message than a more poorly compressed data-set. The best compression should come about when we extract features and use those to recreate the original data. The MDL principle is not used here to actually encode a message, instead our interest is in how long the message would be if we did encode it.

The message length under the MDL formulation, $L(\mathcal{D},\mathcal{H})$, consists of two parts~\cite{mackay2003information}:

\begin{equation}
L(\mathcal{D},\mathcal{H})=L(\mathcal{H}) + L(\mathcal{D}|\mathcal{H})
\label{MDL1}
\end{equation}

\noindent where $L(\mathcal{H})$ is the length of the hypothesis and $L(\mathcal{D}|\mathcal{H})$ encodes the accuracy of the model. $L(\mathcal{H})$ can be interpreted as representing the complexity of the model (or hypothesis), a simple model would have a lower cost (message length) than a more complex model. $L(\mathcal{D}|\mathcal{H})$ then provides a measure of how accurately the model can represent the data, the better the model can reproduce the original data the lower the cost, and the shorter the message required to encode the model errors.

To apply MDL to NMF, we regard the data matrix $\textbf{V}$ as a message to be communicated and the factors, $\textbf{W}$ and $\textbf{H}$, as a reduced representation that can be more efficiently transmitted. The combined number of elements of $\textbf{W}$ and $\textbf{H}$ is usually much smaller than the number of elements in $\textbf{V}$ ($m\times n \gg (m+n)r$) so should, usually, have a shorter required code length. In addition, we usually expect $\textbf{W}$ and $\textbf{H}$ to be fairly sparse compared to $\textbf{V}$ which, again, makes them cheaper to encode, as will be discussed below. We need to reproduce the data matrix $\textbf{V}$ to some pre-agreed precision so we also need a correction matrix, $\textbf{E}=\textbf{V}-\textbf{WH}$ which is the accuracy of the model, with required code length $L(\mathcal{D}|\mathcal{H})$. 

In this MDL formulation the objective function becomes

\begin{align}
f(\textbf{W},\textbf{H})= & L(\mathcal{D},\mathcal{H})\nonumber\\ \nonumber
= & L(\mathcal{H}) + L(\mathcal{D}|\mathcal{H}) \nonumber\\ 
= & L(\textbf{W})+L(\textbf{H})+L(\textbf{E})
\label{eq:objectiveFn}
\end{align}

\noindent where $L(\textbf{W})$, $L(\textbf{H})$ and $L(\textbf{E})$ are the lengths of the code required for $\textbf{W}$, $\textbf{H}$ and $\textbf{E}$ respectively. The terms, $L(\textbf{W})+L(\textbf{H})=L(\mathcal{H})$ can be viewed as a penalty term similar to Tikhonov~\cite{tikhonov1977solutions} and lasso~\cite{tibshirani1996regression} regularisers in classic regression problems. Models of high complexity require correspondingly long codes. However, with a rise in model complexity we expect to see a model which can more accurately recreate the original data matrix and so would expect to see the value of the elements in $\textbf{E}$ fall which should cause a reduction in $L(\textbf{E})$. Therefore, an algorithm that finds a local minimum of Equation~\eqref{eq:objectiveFn} should find a pair of matrices $\textbf{W}$ and $\textbf{H}$ which are automatically regulated to trade-off between complexity and accuracy. 

To understand why smaller errors lead to a shorter length of code consider what happens to a Gaussian distribution as the terms move closer to zero - the distribution becomes tighter. As we need to send the message to a pre-agreed precision, we can see that more terms will be close to zero, with many having the same value, within the pre-agreed precision. These increasing number of terms can be encoded using the same, short, code. The elements far from the Gaussian mean will be more expensive to encode. Increasing accuracy therefore decreases the length of the correction, $L(\mathcal{D}|\mathcal{H})$. If model complexity decreases, for example by an increase in sparsity, we see a similar result, but for the $L(\mathcal{H})$ term - the code required becomes much cheaper as there are many terms near to zero. Therefore, sparse and non-complex model matrices will require a short message to encode them, but will require a longer code for the correction matrix, providing the trade-off between model complexity and accuracy.

In MDL there is no need to know how to encode the matrices, all we need to do is to estimate the length the code would need to be. We do this using the Shannon information content~\cite{shannon1948} which is defined as $h(x)=-\log_{2}P(x)$ where $x$ is the value of an element in the relevant matrix and $P(x)$ is the probability of that value. Including this estimate of the length of the code gives us an objective function:

\begin{align}
f(\textbf{W},\textbf{H})= & -\Bigg[\sum_{i=1}^m\sum_{j=1}^{r}\log_2P(W_{ij}) + \sum_{i=1}^r\sum_{j=1}^{n}\log_2P(H_{ij})\\
&+\sum_{i=1}^m\sum_{j=1}^{n}\log_2P(E_{ij})  \Bigg]
\end{align}

\noindent where $E_{ij}$, $W_{ij}$ and $H_{ij}$ represent the $i^{\text{th}}$, $j^{\text{th}}$ element of $\textbf{E}$, $\textbf{W}$ and $\textbf{H}$ respectively. We have now specified the outline of our model (MDL-NMF) and why it should result in an automatic trade-off between complexity and accuracy. In the next section we will describe our method which estimates the probabilities and then finds appropriate $\textbf{W}$ and $\textbf{H}$ matrices.


\section{Methods and algorithm}
\label{sec:methods}

We propose two methods of estimating the probabilities, in the same manner as the previous work on using MDL to select the subspace size~\cite{squires2017rank}. The first is a non-parametric method of sorting the elements of the matrix into bins and finding the probability by $P(x)=\frac{b_i}{N}$ where $b$ is the number of elements in the $i^{th}$ bin and $N$ is the total number of elements of that matrix. The other, parametric, method applies probability distributions to the $\textbf{E}$, $\textbf{W}$ and $\textbf{H}$ matrices to extract the probability density of each value. 

All the results in this paper were produced using the parametric model as it is computationally faster than the non-parametric approach because the objective function can be differentiated, allowing us to apply gradient descent methods to approach a minimum. Finding a fast non-parametric method might be an interesting future avenue of research.

To find the probability distribution we first put the elements of the $\textbf{W}$, $\textbf{H}$ and $\textbf{E}$ matrices, individually, into bins. We then fit a Gaussian distribution, with mean $\mu$ and standard deviation $\sigma$, to the $\textbf{E}$ matrix and two different gamma distributions to the $\textbf{W}$ (with parameters $\alpha$ and $\beta$) and $\textbf{H}$ (with parameters $a$ and $b$) matrices.

We chose the gamma distribution ($\rho(x)=\frac{\beta^\alpha}{\Gamma(\alpha)}x^{(\alpha-1)}e^{-\beta x}$) because it fulfils several important criteria: it falls to zero as the value becomes large; it remains positive, either falling to zero or rising towards infinity as its elements approach zero; and it is a flexible distribution. While we illustrate the idea here using the gamma and Gaussian distributions, our MDL-NMF method is not dependent on the type of distributions applied, other options may be more suitable for specific applications.

As the objective function of the distribution method is differentiable we seek a solution by following a gradient descent approach. Therefore we need to find $\frac{\partial f}{\partial W_{ij}}$ and $\frac{\partial f}{\partial H_{ij}}$ while noting that $\textbf{E}$ is a function of $\textbf{W}$ and $\textbf{H}$. As with most NMF techniques~\cite{gillis2014and} we update $\textbf{W}$ and $\textbf{H}$ separately, making two separate convex functions to optimise as opposed to the original non-convex problem if we try and update both $\textbf{W}$ and $\textbf{H}$ together. We therefore apply:

\begin{align}
W_{ij}\leftarrow W_{ij}-\lambda_W \frac{\partial f}{\partial W_{ij}}, && H_{ij}\leftarrow H_{ij}- \lambda_H \frac{\partial f}{\partial H_{ij}}
\end{align}

\noindent where $\lambda_W$ and $\lambda_H$ are learning rate parameters, 

\begin{align}
\frac{\partial f}{\partial W_{ij}}=-\frac{[(\alpha-1)W_{ij}^{-1}-\beta ]}{\ln(2)}-\frac{1}{\ln(2)\sigma^2}\sum_{k=1}^{n}\left[(E_{ik}-\mu)H_{jk}\right] 
\end{align}

\noindent and

\begin{align}
\frac{\partial f}{\partial H_{ij}}=-\frac{[(a-1)H_{ij}^{-1}-b ]}{\ln(2)}-\frac{1}{\ln(2)\sigma^2}\sum_{k=1}^{m}\left[(E_{kj}-\mu)W_{ki}\right]. 
\end{align}

We can also consider the updates from a matrix multiplication perspective which can significantly speed up computational time. In this case we have:

\begin{align}
\boldsymbol{W}\leftarrow \boldsymbol{W}-\lambda_W \nabla_{\boldsymbol{W}}f, && \boldsymbol{H}\leftarrow \boldsymbol{H}-\lambda_H \nabla_{\boldsymbol{H}}f
\end{align}

where 

\begin{align}
\nabla_{\boldsymbol{W}}f = \boldsymbol{\widetilde{W}}+\boldsymbol{\widetilde{E}}\boldsymbol{H}^T && \nabla_{\textbf{H}}f = \boldsymbol{\widetilde{H}}+\boldsymbol{W}^T\boldsymbol{\widetilde{E}}
\end{align}

and we have defined three new matrices, $\boldsymbol{\widetilde{W}}$, $\boldsymbol{\widetilde{H}}$ and $\boldsymbol{\widetilde{E}}$, with elements: 

\begin{align}
&\widetilde{W}_{ij}=-\Big( \frac{(\alpha-1)W_{ij}^{-1}-\beta}{\ln(2)}\Big), \widetilde{H}_{ij}=-\Big( \frac{(a-1)H_{ij}^{-1}-b}{\ln(2)}\Big), \\ 
&\widetilde{E}_{ij}=\frac{1}{\ln(2)\sigma^2}(E_{ij}-\mu)
\end{align}

\noindent respectively.

As the $\textbf{W}$ and $\textbf{H}$ matrices are changed, the distributions will also shift, so we also need to change the parameters of the distributions in response. In Algorithm~\ref{al:MDLObjFn-distAlgorithm} we specify the algorithm for finding the $\textbf{W}$ and $\textbf{H}$ matrices for the MDL-NMF formulation. The precision term $\delta$ depends on the data and sets the bin width. We do not impose a specific stopping criteria - generally we run the algorithm until there appears to be no further fall in the objective function or it begins to rise.

\begin{algorithm}
\caption{MDL-NMF implementation}
\label{al:MDLObjFn-distAlgorithm}
\begin{algorithmic}[1]
\Statex \textbf{Input}: Data matrix $\textbf{V}$, subspace size $r$.
\Statex \textbf{Output}: $\textbf{W}$, $\textbf{H}$
\State \textbf{Initialise}: $\textbf{W}^{(o)}$, $\textbf{H}^{(o)}$.
\State \textbf{Define}: learning rate $\lambda$, precision $\delta$
\State \textbf{Calculate}: parameters of the distribution, initial objective function $f_0$
\State \textbf{for} t=1,2,....stopping criteria reached
\State \hspace{1cm}\textbf{if} $W_{i,j}<\delta/2$ or $H_{i,j}<\delta/2$ set to $\delta/2$
\State \hspace{1cm}Update: $\textbf{W}\leftarrow\textbf{W}-\lambda\nabla_{\textbf{W}}f_{t-1}$
\State \hspace{1cm}Update: $\textbf{H}\leftarrow\textbf{H}-\lambda\nabla_{\textbf{H}}f_{t-1}$
\State \hspace{1cm}Calculate $f_t$
\State \hspace{1cm}\textbf{if} $f_t\ge f_{t-1}$ reduce $\lambda$, revert $\textbf{W}$ and $\textbf{H}$ to previous values
\State \hspace{1cm}Recalculate: parameters of distributions using maximum likelihood estimates.
\State \textbf{end for}

\end{algorithmic}
\end{algorithm}

Perhaps the most widely used NMF algorithm is the multiplicative updates (MU) method of Lee and Seung~\cite{lee1999learning,lee2001algorithms} which is, effectively, gradient descent of the Frobenius norm error with an automatically changing learning rate which ensures that the objective function monotonically decreases. In contrast, due to the static learning rate, we have no guarantee that the MDL-NMF objective function will monotonically decrease. We compensate for this by reducing the learning rate if the objective function is not falling.

If different distributions were chosen the only change required would be in the $\nabla_{\boldsymbol{W}}f$ and $\nabla_{\boldsymbol{H}}f$ terms. These would need to be worked out again for the new distributions, otherwise the method and algorithm would remain the same.


\section{Results and Discussion}
\label{sec:results}
In the previous sections we have given an explanation of why our MDL-NMF method should work and the potential advantages over methods that minimise an error alone, or those that impose sparsity with ad-hoc parameter tuning. We now present empirical results to demonstrate the efficacy of our approach. In Table~\ref{tab:table1} we display the three data-sets we tested our MDL-NMF method on, including the type of data, the number of dimensions and number of data-points. These data-sets were chosen for their considerable heterogeneity to show the breadth of application of our method.

\begin{table}[ht]
\caption{Data-sets, the type of data, the number of dimensions, m, and number of data-points, n.}
\label{tab:table1}
\begin{center}
\begin{tabular}{ l l l l l }
\hline
\textbf{Name} & \textbf{Type} & \textit{m} & \textit{n} & \textbf{Source} \\
\hline
Faces         	& Image         & 361        & 2429     & http://cbcl.mit.edu/\\
				&				&			 &			& software-datasets \\
				&				&			 &			& /FaceData2.html \\
Transcriptome	& Biological    & 5000       & 38       & http://www.broad \\
				&				&			 &			& institute.org/cgi-bin \\
				&				&			 &			& /cancer/datasets.cgi \\
FTSE 100       	& Financial     & 1305       & 94       & University Bloomberg \\  
				&				&			 &			& information terminal \\
\hline
\end{tabular}
\end{center}
\end{table}

The faces data is a set of 19-by-19 grey-scale images of faces which are individually converted to a 361 dimension vector then stacked up to make the $361-\text{by}-2429$ matrix. The transcriptome~\cite{golub1999molecular} data set is gene expression data for 38 samples, for people with two different types of leukemia. Each column of the FTSE 100 data represents the day closing price of one stock over a five year period, these are stacked with the other companies (94 rather than than 100 due to companies dropping into and out of the FTSE 100 over time) from the FTSE 100 to make our data-set. These data-sets are from a wide range of applications, all have previously been studied using NMF techniques,~\cite{lee1999learning,devarajan2006nonnegative,squires2017non}, and provide evidence for the large range of effective use of MDL-NMF. They are from very different domains, with significantly different $m/n$ ratios.

Making a direct comparison between dimensionality reduction techniques is a challenge. Unlike in supervised learning there is no direct model to test held-out data on, therefore we cannot easily check the generalisation error to compare our results to other methods. To attempt to compensate for this we have also tested MDL-NMF on semi-synthetic data.

The use of synthetic data has two main problems: 1) that it is difficult to know what the underlying structure should look like; 2) that synthetic data can be created and adjusted to provide data that produce favourable results for any particular technique instead of a realistic portrayal of the effectiveness of the method. Therefore, instead of using completely synthetic data we used a form of semi-synthetic data based upon the three data-sets in Table~\ref{tab:table1}. 

Our general principle to create semi-synthetic data is to perform NMF on the real data, producing $\textbf{W}$ and $\textbf{H}$. From there we know what the underlying structure of the semi-synthetic data is - assuming standard NMF uncovers real structure. We then rebuild the data, add noise and use our MDL-NMF technique to attempt to uncover the known true results. 

We have not provided extensive results for how long the algorithm takes to run - if using the matrix multiplication method a full run from a randomised starting point to final solution for the largest and slowest dataset (faces) takes less than three minutes on an ordinary desktop computer. We would suggest the MDL-NMF method may be better used as a tuning method, with initial $\textbf{W}$ and $\textbf{H}$ matrices set as similar to the results from standard NMF, in which case MDL-NMF runs even faster. Either way MDL-NMF, while slower than standard NMF, takes an insignificant length of time to run on these data-sets. The scalability of this work to much larger datasets would require further research.

In the following subsections we demonstrate the potential value of MDL-NMF by investigating: 1) how well it reproduces the original data; 2) the process by which MDL-NMF learns the representation; 3) the ability of MDL-NMF to extract signal rather than noise from the semi-synthetic data.


\subsection{Recreation of the original data}

Any NMF algorithm must be able to reproduce the original data with a reasonable degree of similarity. The precise level of similarity desired is not straight-forward to specify because part of the purpose of NMF is to suppress noise in the data. In any real data set we do not know what is signal and what is noise. Therefore, if an NMF algorithm perfectly recreates a real world data-set we would not necessarily consider that a success, as it is likely to be modelling noise. On the other hand, if we are using data-sets with a reasonably high signal to noise ratio we would expect to be able to recreate much of the data using NMF techniques.

\begin{figure}
\centering
\subfigure[]{
\includegraphics[scale=0.35]{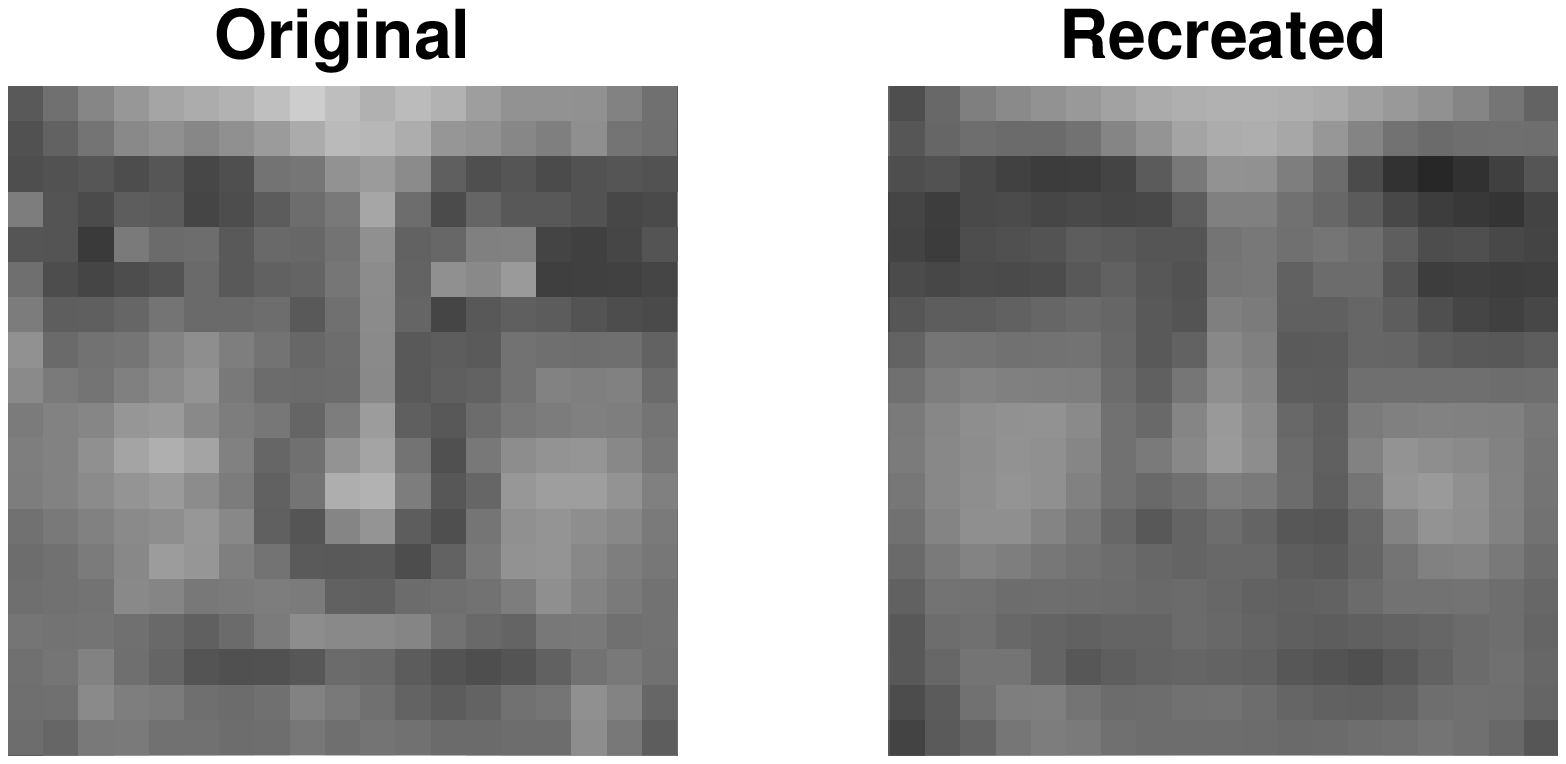}
\label{fig:recreation1a}
}
\subfigure[]{
\includegraphics[scale=0.35]{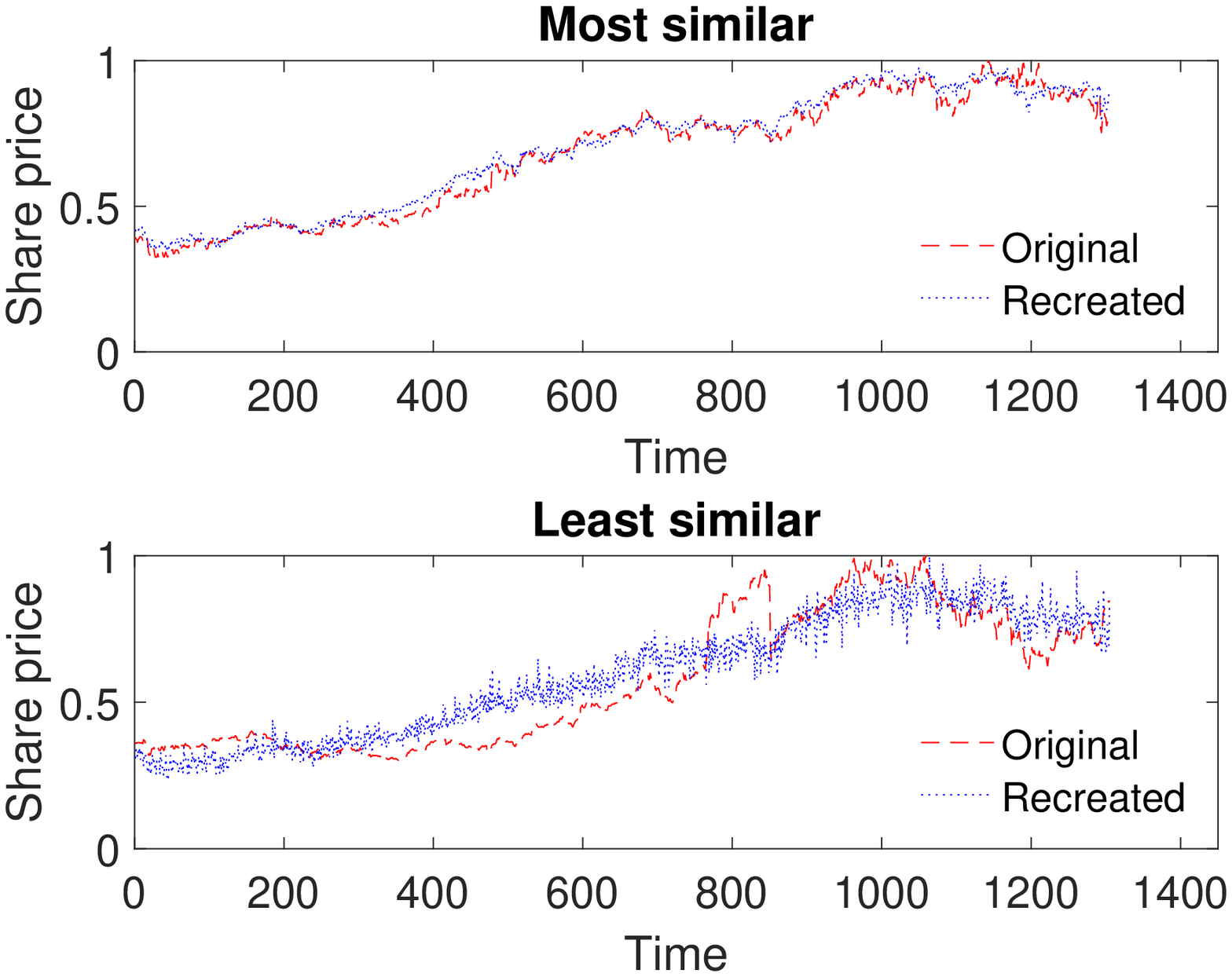}
\label{fig:recreation1b}
}
\subfigure[]{
\includegraphics[scale=0.35]{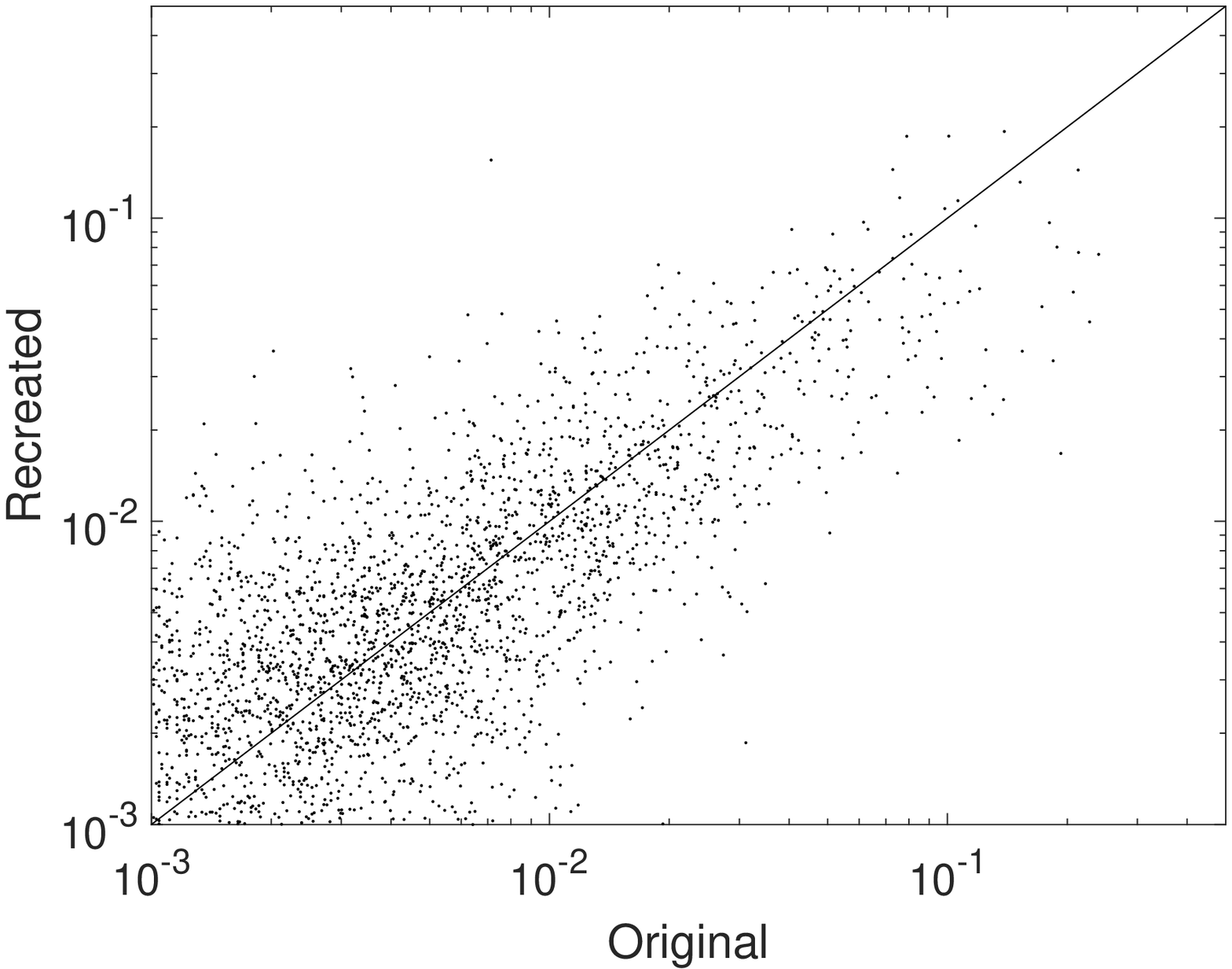}
\label{fig:recreation1c}
}
\caption{Plots showing the quality of the final recreated data. (a) One randomly selected sample from the images of faces dataset. The left image is the original and the right image is the recreated face for $r=80$. (b) Actual (red dashed line) and recreated (blue dotted line) prices against time for one stock for $r=10$. The top is the stock with most similarity between actual and recreated, while the bottom plot is the least similar. (c) 5000 randomly selected elements of the data matrices with the recreated value plotted against original values for the transcriptome data with $r=4$. MDL-NMF effectively recreates all three datasets.}
\label{fig:recreation1}
\end{figure}

In Figure~\ref{fig:recreation1} examples of the final recreated output from each of the three data-sets are shown. In Figure~\ref{fig:recreation1a} we show one randomly selected sample from the images of faces data-set. The left face is the original with the recreated face (for $r=80$) on the right. In Figure~\ref{fig:recreation1b} are plots of original (red dashed line) and recreated (blue dotted line) prices against time for one stock from the FTSE 100 dataset (with $r=10$). The top and bottom plots show the most and least accurately recreated stocks respectively. The stock that is recreated most inaccurately, still produces a good reproduction of the actual prices. In Figure~\ref{fig:recreation1c}, for ease of visualisation of the genomic data, we randomly sampled 5000 of the elements of the data matrix $\textbf{V}$ (x-axis) and compared the values of them to the same elements of the recreated matrix $\textbf{WH}$ (y-axis) with $r=4$. For all three data-sets our MDL-NMF method produces an effective recreation of the data.

\subsection{The learning process}

\begin{figure}
\centering
\subfigure[]{
\includegraphics[scale=0.45]{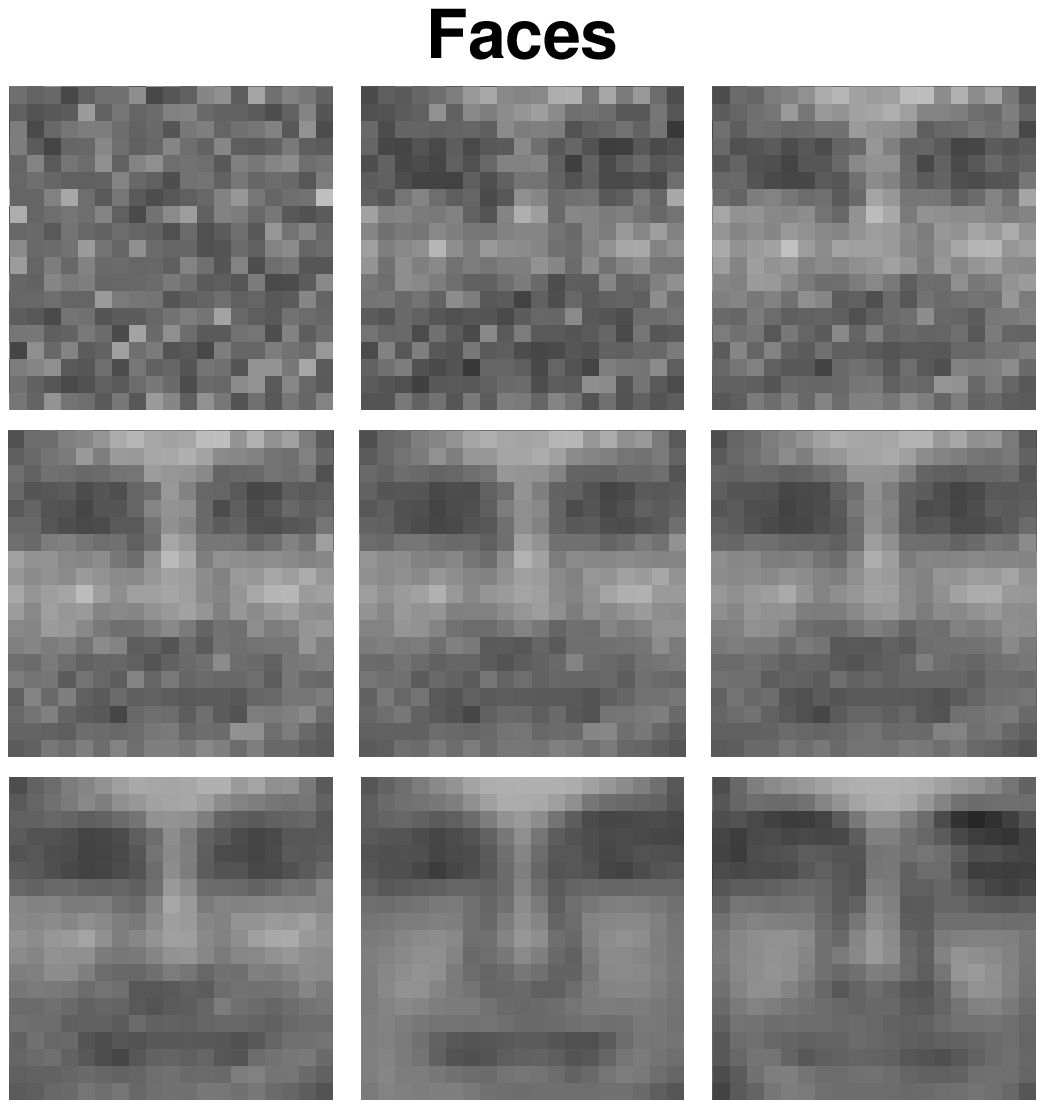}
\label{fig:learningProcess1a}
}
\subfigure[]{
\includegraphics[scale=0.45]{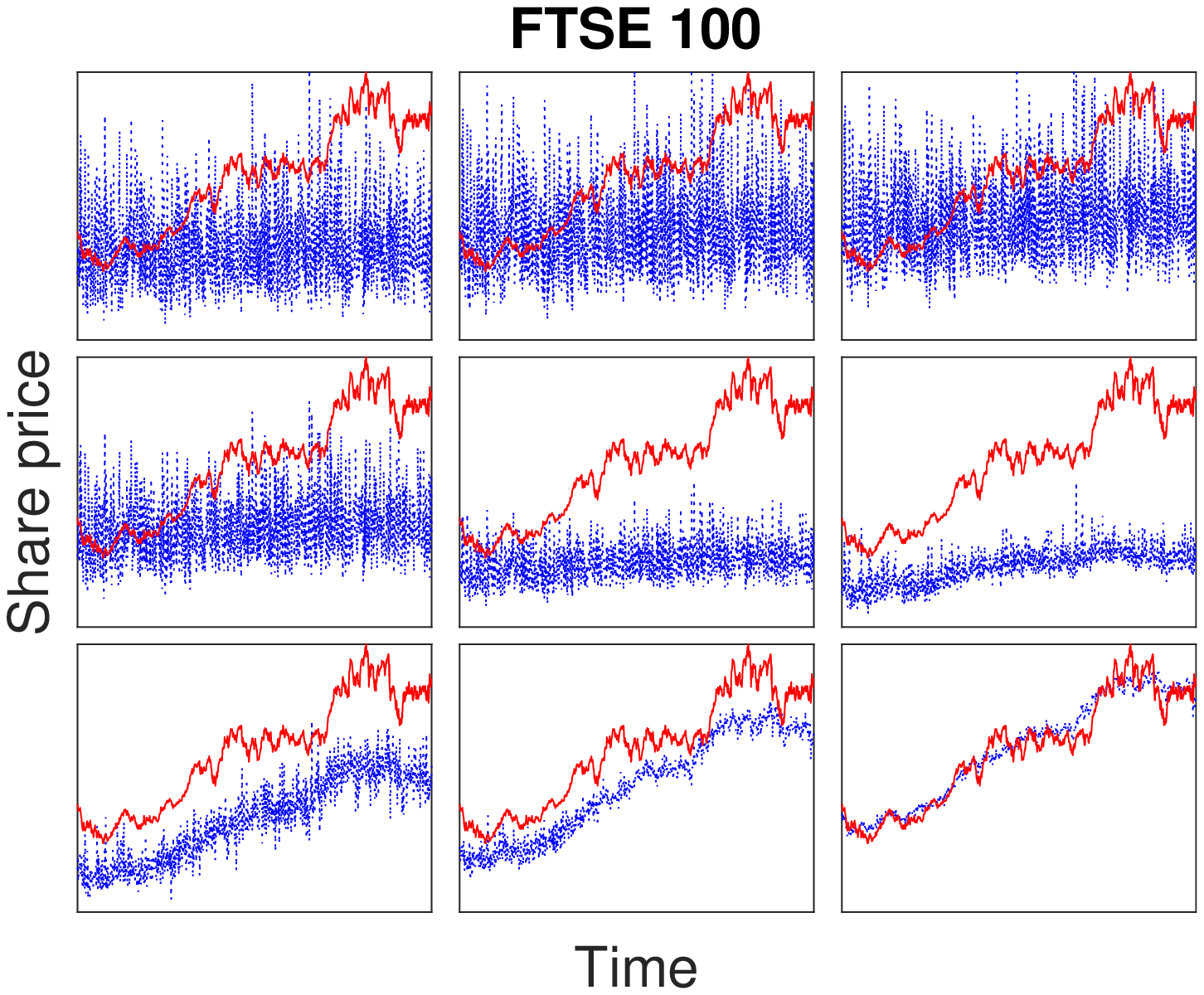}
\label{fig:learningProcess1b}
}
\subfigure[]{
\includegraphics[scale=0.45]{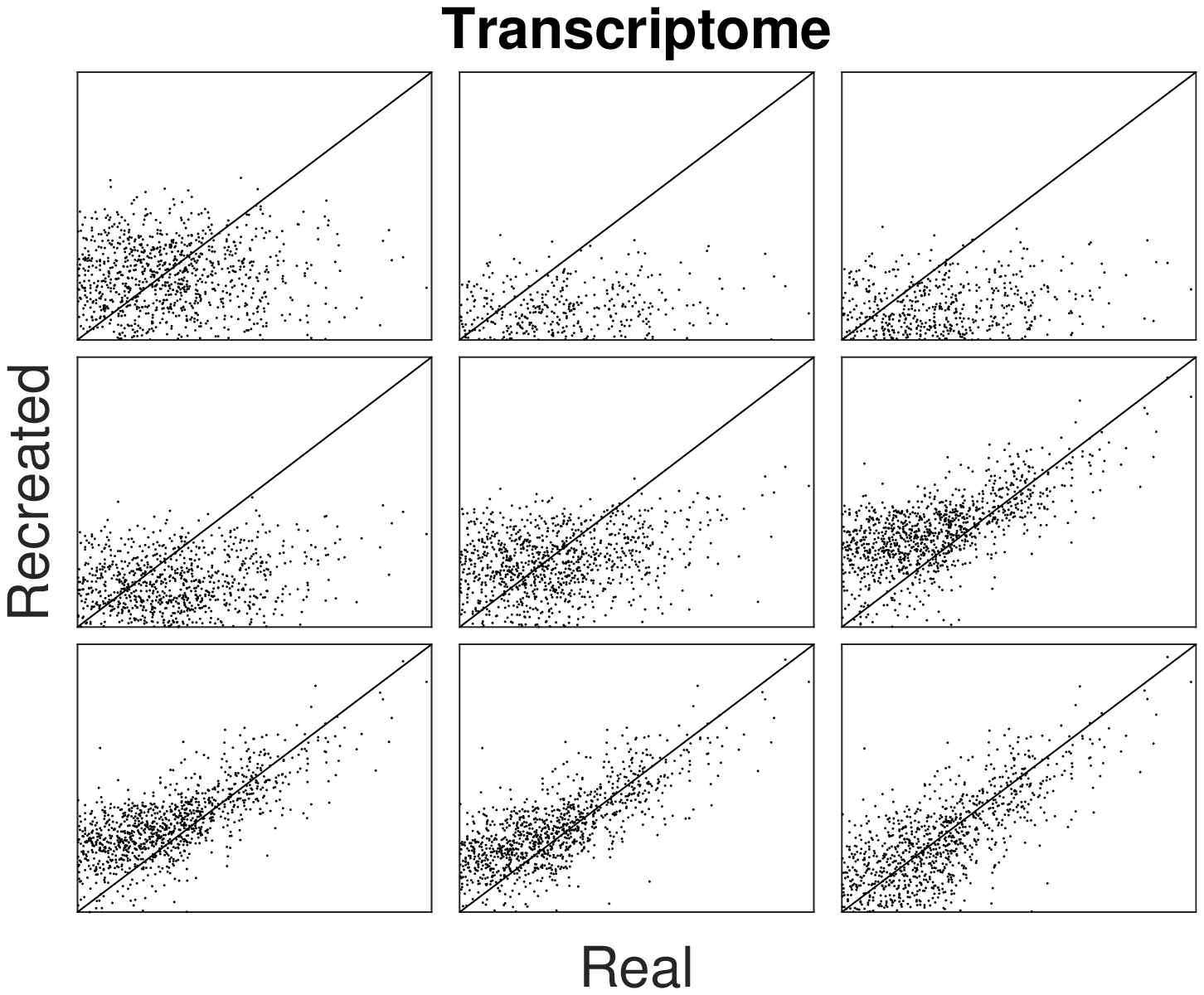}
\label{fig:learningProcess1c}
}
\caption{The reproduced data recorded at different iterations of the training process. (a) A sample from the faces dataset; (b) a randomly selected stock with the real (red dashed) and recreated (blue dotted) prices; (c) 5000 recreated results versus original values for the transcriptome data. All three show the change from a randomised starting point to a reasonable recreation of the original data.}
\label{fig:learningProcess1}
\end{figure}

We now observe the learning dynamics by taking snapshots of the reconstruction during the learning process. Our MDL-NMF objective function is not purely based on minimising the error and therefore may follow a somewhat different trajectory to finding a minimum, it is therefore useful to demonstrate what does happen as we reduce the objective function. In Figure~\ref{fig:learningProcess1} we show how the reproduced data looks as the algorithm proceeds. Due to the differences between the data-sets they take a different number of iterations to arrive at a reasonable solution and the snapshots were chosen at appropriate iteration intervals to demonstrate how the quality of the fit improves. Figure~\ref{fig:learningProcess1a} is an image from the faces dataset; Figure~\ref{fig:learningProcess1b} is a FTSE 100 stock with the real (red dashed) and recreated (blue dotted) prices changing with iterations; Figure~\ref{fig:learningProcess1c} is 5000 of the elements of the recreated matrix versus the real values from the transcriptome data. All three plots show the clear trend from a randomised starting point to a reasonable final recreation of the original data.

\begin{figure}
\centering
\subfigure[]{
\includegraphics[scale=0.45]{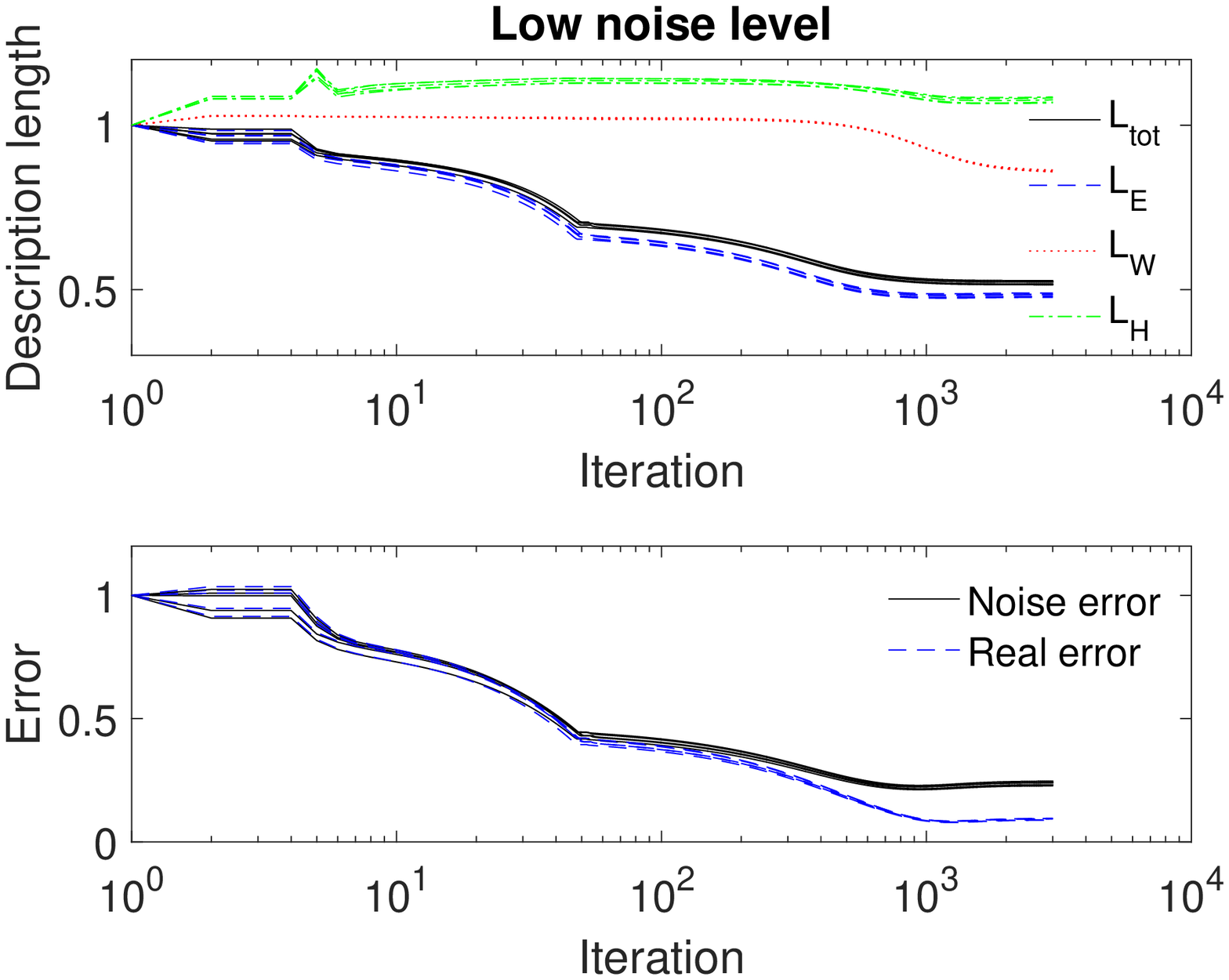}
\label{fig:synData1a}
}
\subfigure[]{
\includegraphics[scale=0.45]{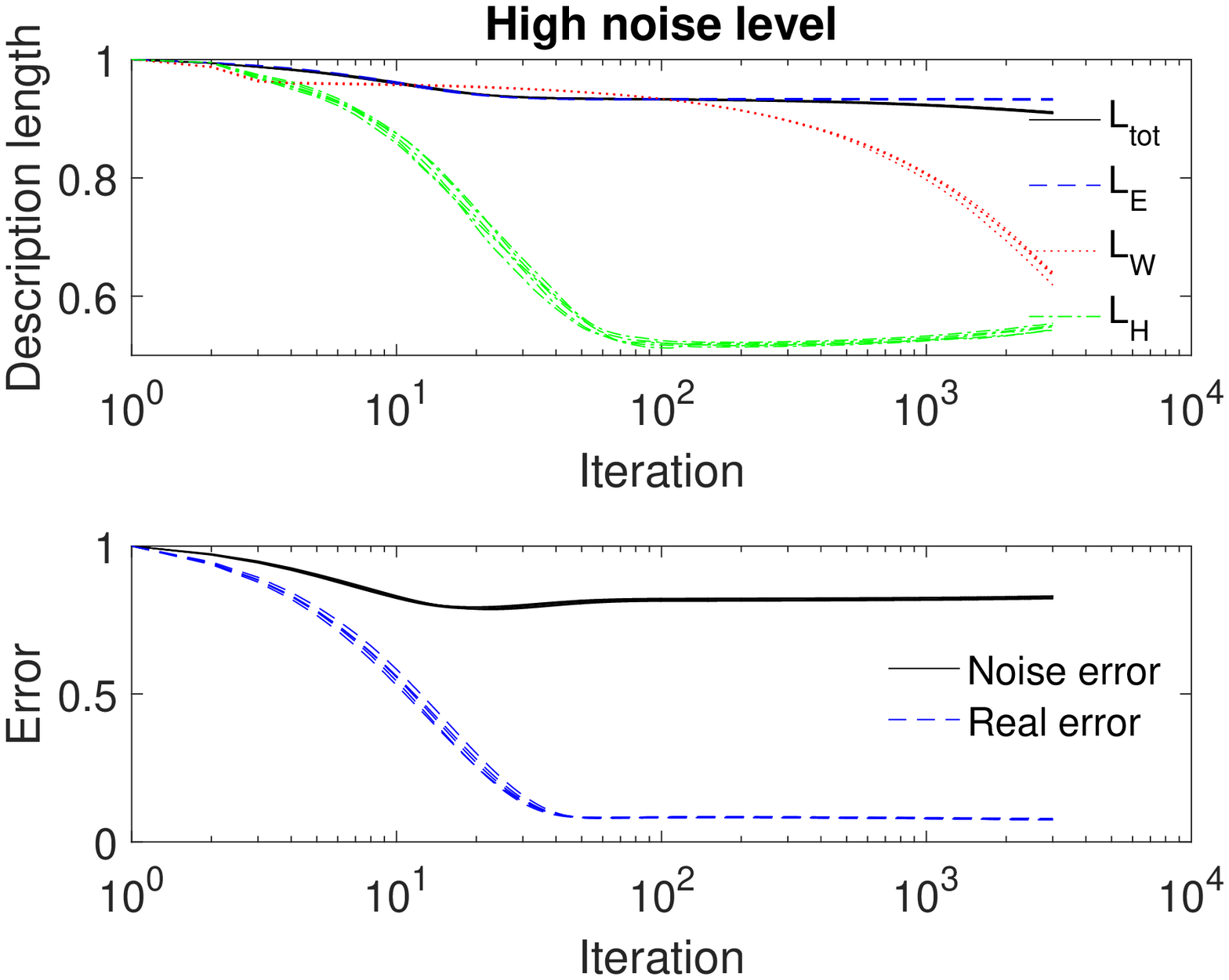}
\label{fig:synData1b}
}
\caption{The changes in description lengths and errors with iteration for the semi-synthetic transcriptome data for a low noise variant and a high noise variant.}
\label{fig:synData1}
\end{figure}

In Figure~\ref{fig:synData1} we display the changes in both the description lengths and errors with iteration for a low and high noise variant of the semi-synthetic transcriptome data. There are five repeats of the MDL-NMF algorithm displayed on the same graph for the same data with different initialisations of the $\textbf{W}$ and $\textbf{H}$ matrices. For these plots we initialised the matrices for MDL-NMF by running standard NMF on the semi-synthetic data and then adding Gaussian noise to each element of the two matrices. The true error is defined as $\textbf{E}_{\text{true}}=||\textbf{V}_{\text{true}}-\textbf{WH}||_{\text{Fro}}^2$ where $\textbf{V}_{\text{true}}$ is the underlying data matrix before we added noise while the noisy error is $\textbf{E}_{\text{noise}}=||\textbf{V}_{\text{noise}}-\textbf{WH}||_{\text{Fro}}^2$ where $\textbf{V}_{\text{noise}}$ is the semi-synthetic data-set after noise is added. All the plots are normalised so the description lengths and errors all start at one.

There are some interesting features of the graphs. At a high noise variant the MDL-NMF significantly reduces the true error while the noisy error is not reduced much. At the low noise level most of the reduction in description length is achieved by reducing the noisy error - that also reduces the true error. With the high noise level the $L_E$ terms does not reduce much, instead we get a much larger relative reduction in the $L_W$ and $L_H$ values with commensurate significant falls in the true error. This may imply that the algorithm is reducing the complexity of the model ($L_W$ and $L_H$) in preference to the error term ($L_E$) when there is a high level of noise. These type of plots may be useful analysis tools to use when investigating new data sets.


\subsection{Representing true data over noise}

A good NMF technique should be able to extract true data from noisy data. For our semi-synthetic data we can look at how well our MDL-NMF technique does with extracting the true data from the noise we added.

\begin{figure}
\centering
\subfigure[]{
\includegraphics[scale=0.42]{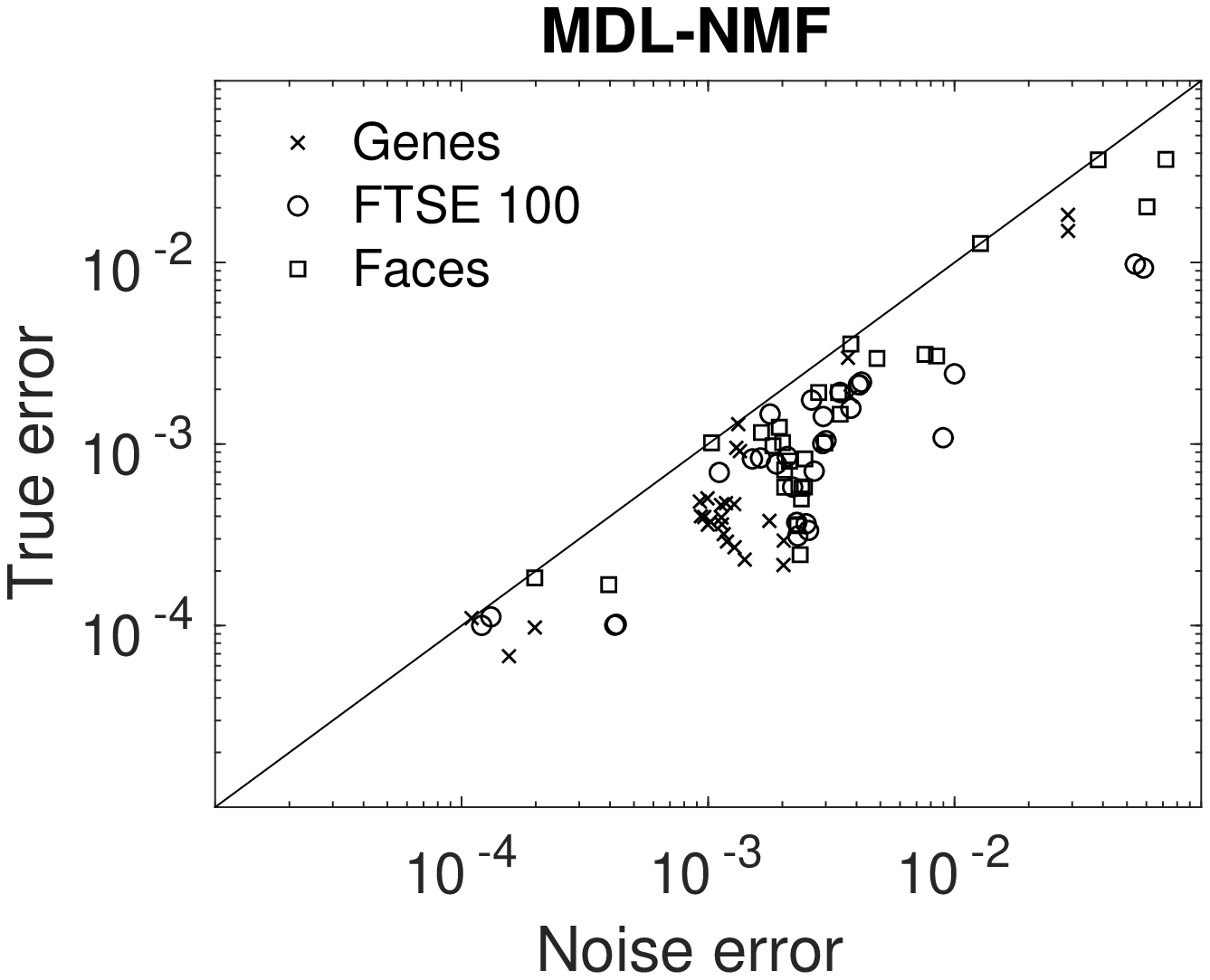}
\label{fig:synData2a}
}
\subfigure[]{
\includegraphics[scale=0.42]{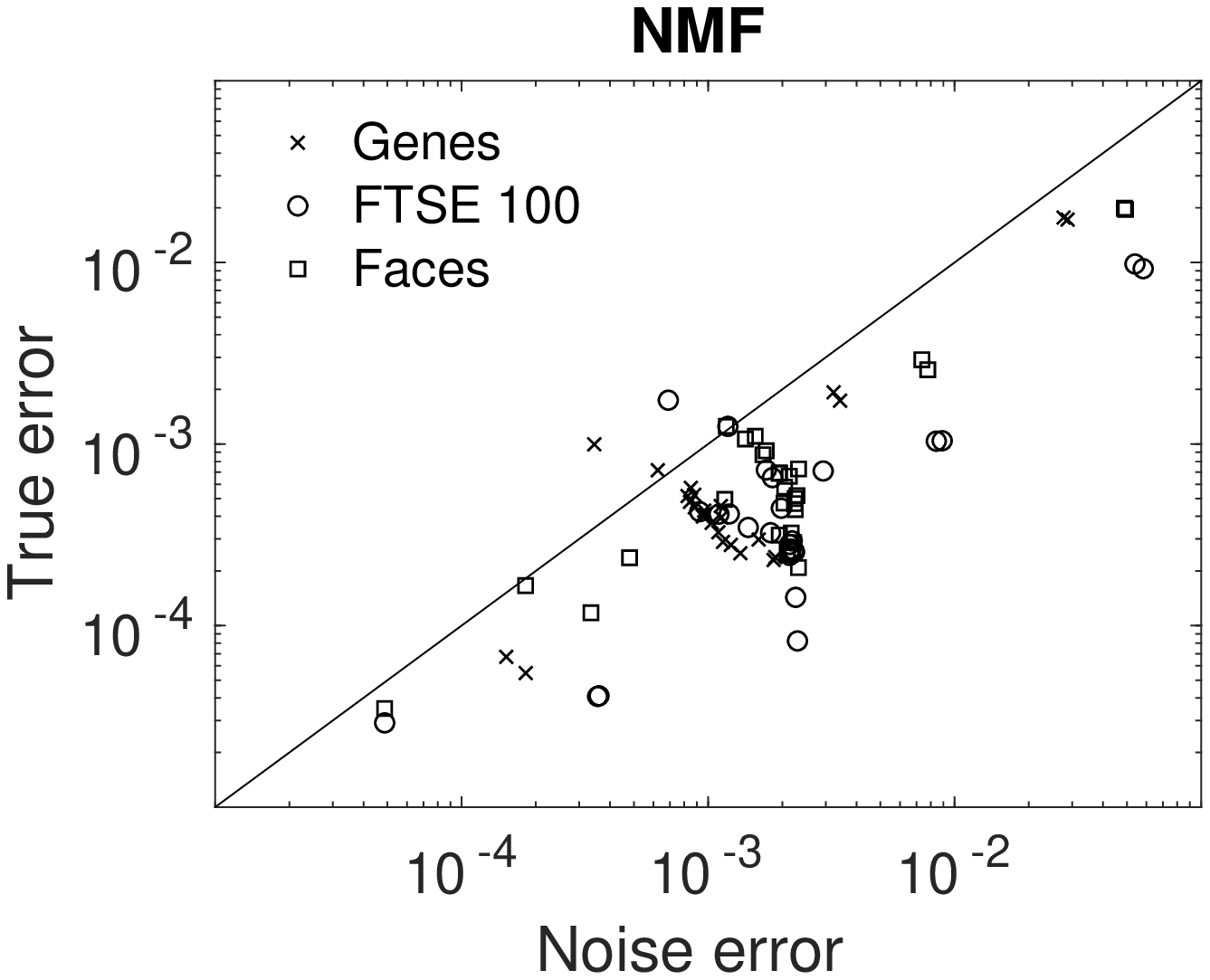}
\label{fig:synData2b}
}
\caption{The real error against noise error show that MDL-NMF almost always finds a better fit to the real data than the noise added data. Results for a range of semi-synthetic data-set types are shown. (a) for MDL-NMF. (b) for NMF.}
\label{fig:synData2}
\end{figure}

In Figure~\ref{fig:synData2a} we show the true error against the noise error for MDL-NMF. Any point below the straight line shows a version where MDL-NMF finds a better error on the true data than the noise data. These results are from a range of variants of the semi-synthetic data: varying additional noise levels; varying $r$ values used to create the semi-synthetic data; varying $r$ values used to try and extract the true representations; true $\textbf{W}$ and $\textbf{H}$ matrices that are smoothed; using sparsity induced NMF (sNMF) to create the true matrices. The different symbols represent the three different types of semi-synthetic data. It is clear that almost all the results show a lower true error than noisy error MDL-NMF is able to cut through the added noise to represent the true data better than the noise data. 

In Figure~\ref{fig:synData2}b we show the same results but for standard NMF using multiplicative updates~\cite{lee1999learning}. We note here that we originally made the data using a NMF method - we might therefore expect NMF to effectively find the real data beneath - which it effectively does for most of the data points but there are several points where standard NMF produces a better fit to the noisy data than the true data, it is overfitting to the noise.

A direct comparison between NMF and MDL-NMF may not be completely fair, as NMF was used to create the semi-synthetic data in the first place. Also we are assuming that the original creation of the semi-synthetic data removed all the noise from the original data - if it did not then there is still noise left in the true error. Accepting these limitations, if we see any improvement in representing real data using MDL-NMF over standard NMF then it implies that MDL-NMF can be a useful tool to use alongside NMF.

\begin{figure}
\centering
\includegraphics[scale=0.45]{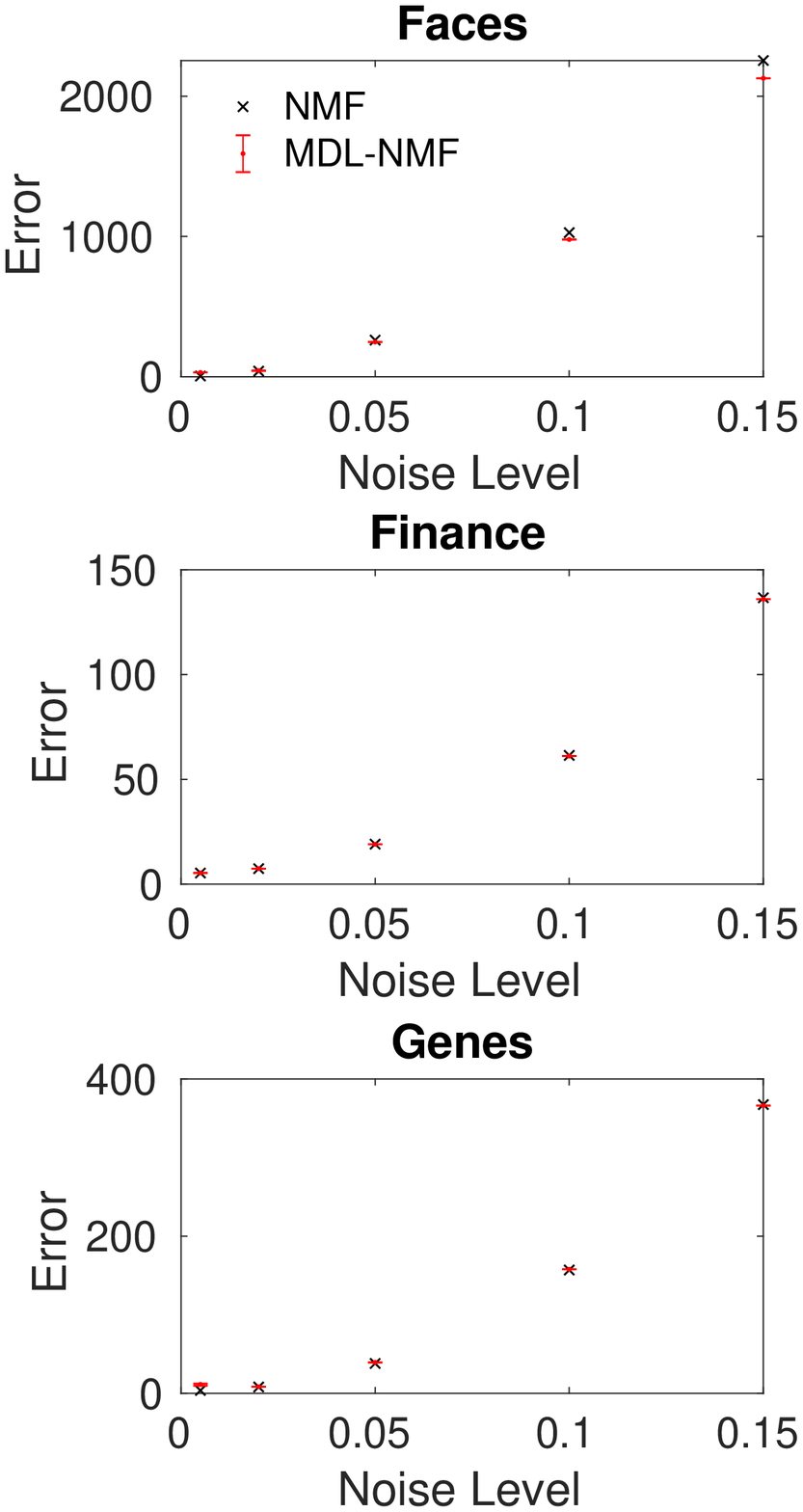}
\includegraphics[scale=0.45]{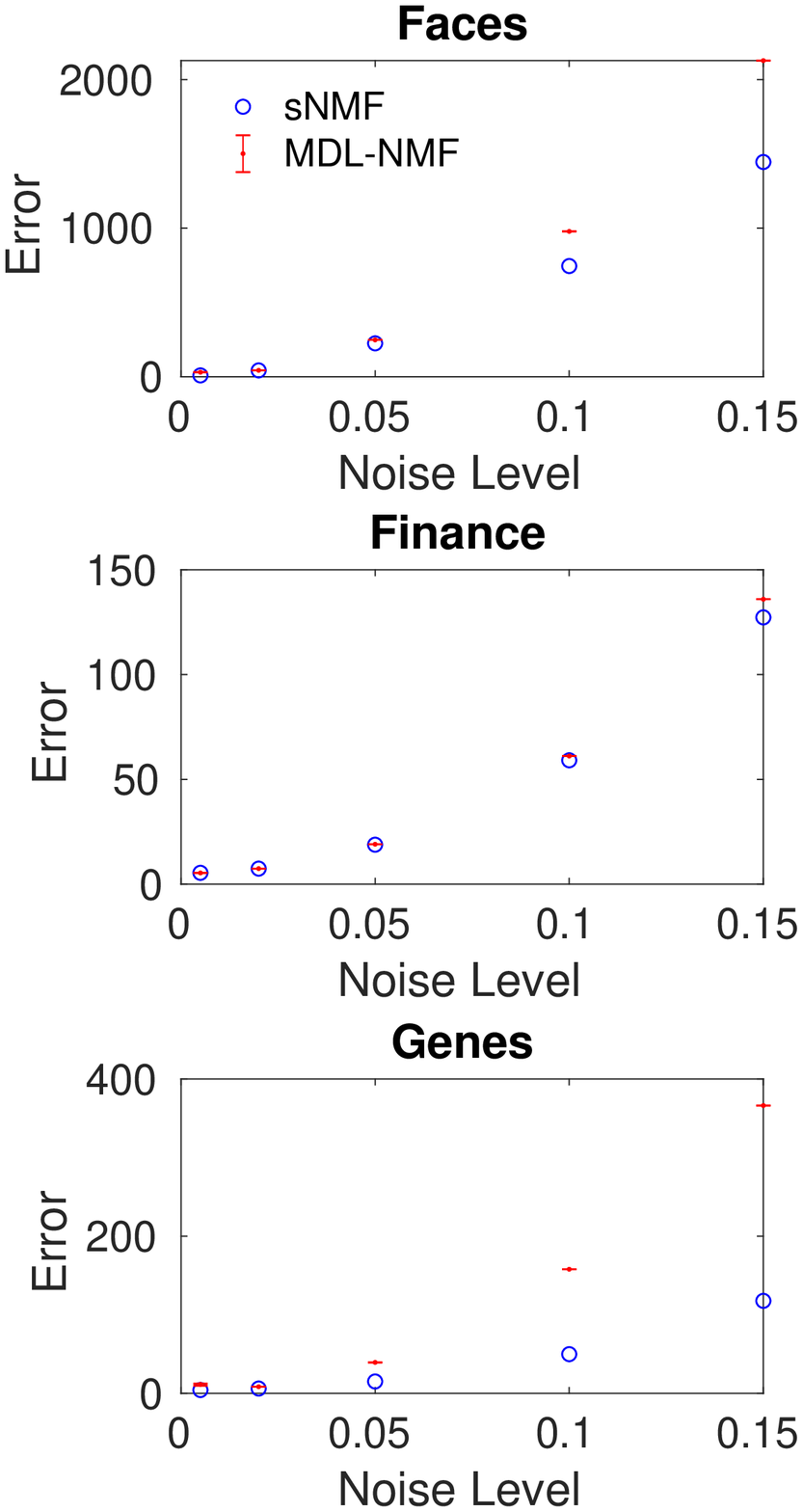}
\caption{A comparison of the real error found for MDL-NMF against NMF (three plots on the left) and sNMF (three plots on the right) for the three semi-synthetic data sets.}
\label{fig:synData3}
\end{figure}

In Figure~\ref{fig:synData3}a we show real error against levels of noise for the three semi-synthetic datasets for both MDL-NMF and NMF. In Figure~\ref{fig:synData3}b we display the same but this time using sparsity induced NMF (sNMF) from Hoyer~\cite{hoyer2004non} with induced sparsity chosen to be the actual sparsity of the known semi-synthetic $\textbf{W}$ and $\textbf{H}$. This favours the sNMF over MDl-NMF which has no direct information about the level of sparsity of the $\textbf{W}$ and $\textbf{H}$ matrices, we would expect sNMF to produce significantly better results as the noise increases. 

The comparison between MDL-NMF and NMF shows very little difference between the two methods with low level of noise, but as the noise level increases we get an improved representation with MDL-NMF. This is true for all three data-sets although it is most apparent for the Faces. This finding is sensible, at low levels of additional noise there is little need to regularise but as the noise level increases we would expect to see increased levels of over-fitting. The comparison with sNMF is instructive on this, when we set the sparseness level to what the underlying matrices have, we get much less overfitting - the sNMF does much better than either MDL-NMF and NMF. So the sparseness is reducing much of the overfitting. MDL-NMF shows a promising ability to, at least a little, compensate for the increases in noise. We initialised the MDL-NMF matrices using final NMF matrices with a little noise added, so it is possible that with better initialisation choices we could do even better than the small improvements we show here.

\section{Conclusions}
\label{sec:conclusions}
The standard NMF objective function which minimizes the error may result in producing matrices which are too complex and overfit the data. We provide an alternative, based upon an MDL approach which provides an automatic and natural tradeoff between the accuracy and complexity of the representation. 

We demonstrated the effectiveness of MDL-NMF on three heterogeneous data-sets, producing representations which match well with the real data. We also tested our model on semi-synthetic data producing representations which represent the real data significantly more than the noise. In addition, studying the changes in the description lengths with iteration may be a useful analysis tool to use to investigate a dataset.

MDL-NMF produces superior results at extracting real data from noise over NMF for some semi-synthetic data especially when the added noise becomes high. It would be a worthwhile procedure to run MDL-NMF after a standard NMF run using the NMF matrices as initialisations for MDL-NMF to see how the two representations compare. 

We have proposed using an MDL based objective function for NMF and demonstrated that it works effectively. However, the details are highly flexible. The choice of both distributions to model the matrices and the optimization scheme for finding the $\textbf{W}$ and $\textbf{H}$ matrices may both be subject to considerable improvement. There may also be significantly faster and more efficient methods of finding the $\textbf{W}$ and $\textbf{H}$ matrices in the MDL-NMF formulation than that proposed in this paper. The choice of the learning rates $\lambda_W$ and $\lambda_H$ could potentially be improved as could the initialisation of the $\textbf{W}$ and $\textbf{H}$ matrices. Our aim is to introduce the method and background and demonstrate that it has potential to be a useful tool.

\section*{References}

\bibliography{refs}

\begin{thebibliography}{10}
\expandafter\ifx\csname url\endcsname\relax
  \def\url#1{\texttt{#1}}\fi
\expandafter\ifx\csname urlprefix\endcsname\relax\def\urlprefix{URL }\fi
\expandafter\ifx\csname href\endcsname\relax
  \def\href#1#2{#2} \def\path#1{#1}\fi

\bibitem{devarajan2006nonnegative}
K.~Devarajan, Nonnegative matrix factorization—a new paradigm for large-scale
  biological data analysis, in: Proceedings of the Joint Statistical Meetings.
  Joint Statistical Meetings, 2006, pp. 6--10.

\bibitem{lee1999learning}
D.~D. Lee, H.~S. Seung, Learning the parts of objects by non-negative matrix
  factorization, Nature 401~(6755) (1999) 788--791.

\bibitem{wang2011community}
F.~Wang, T.~Li, X.~Wang, S.~Zhu, C.~Ding, Community discovery using nonnegative
  matrix factorization, Data Mining and Knowledge Discovery 22~(3) (2011)
  493--521.

\bibitem{lee2001algorithms}
D.~D. Lee, H.~S. Seung, Algorithms for non-negative matrix factorization, in:
  Advances in neural information processing systems, 2001, pp. 556--562.

\bibitem{paatero1994positive}
P.~Paatero, U.~Tapper, Positive matrix factorization: A non-negative factor
  model with optimal utilization of error estimates of data values,
  Environmetrics 5~(2) (1994) 111--126.

\bibitem{hoyer2004non}
P.~O. Hoyer, Non-negative matrix factorization with sparseness constraints, The
  Journal of Machine Learning Research 5 (2004) 1457--1469.

\bibitem{squires2017rank}
S.~Squires, A.~Pr{\"u}gel-Bennett, M.~Niranjan, Rank selection in nonnegative
  matrix factorization using minimum description length, Neural Computation.

\bibitem{schmidt2009bayesian}
M.~N. Schmidt, O.~Winther, L.~K. Hansen, Bayesian non-negative matrix
  factorization, in: International Conference on Independent Component Analysis
  and Signal Separation, Springer, 2009, pp. 540--547.

\bibitem{blei2010bayesian}
D.~M. Blei, P.~R. Cook, M.~Hoffman, Bayesian nonparametric matrix factorization
  for recorded music, in: Proceedings of the 27th International Conference on
  Machine Learning (ICML-10), 2010, pp. 439--446.

\bibitem{wang2013online}
N.~Wang, J.~Wang, D.-Y. Yeung, Online robust non-negative dictionary learning
  for visual tracking, in: Proceedings of the IEEE International Conference on
  Computer Vision, 2013, pp. 657--664.

\bibitem{rissanen1978modeling}
J.~Rissanen, Modeling by shortest data description, Automatica 14~(5) (1978)
  465--471.

\bibitem{wallace1968information}
C.~S. Wallace, D.~M. Boulton, An information measure for classification, The
  Computer Journal 11~(2) (1968) 185--194.

\bibitem{barron1998minimum}
A.~Barron, J.~Rissanen, B.~Yu, The minimum description length principle in
  coding and modeling, Information Theory, IEEE Transactions on 44~(6) (1998)
  2743--2760.

\bibitem{mackay2003information}
D.~J. MacKay, Information theory, inference and learning algorithms, Cambridge
  university press, 2003.

\bibitem{tikhonov1977solutions}
A.~N. Tikhonov, V.~I. Arsenin, F.~John, Solutions of ill-posed problems,
  Vol.~14, Winston Washington, DC, 1977.

\bibitem{tibshirani1996regression}
R.~Tibshirani, Regression shrinkage and selection via the lasso, Journal of the
  Royal Statistical Society. Series B (Methodological) (1996) 267--288.

\bibitem{shannon1948}
C.~E. Shannon, A mathematical theory of communication, The Bell System
  Technical Journal 27~(3) (1948) 379--423.
\newblock \href {https://doi.org/10.1002/j.1538-7305.1948.tb01338.x}
  {\path{doi:10.1002/j.1538-7305.1948.tb01338.x}}.

\bibitem{gillis2014and}
N.~Gillis, The why and how of nonnegative matrix factorization, Regularization,
  Optimization, Kernels, and Support Vector Machines 12 (2014) 257.

\bibitem{golub1999molecular}
T.~R. Golub, D.~K. Slonim, P.~Tamayo, C.~Huard, M.~Gaasenbeek, J.~P. Mesirov,
  H.~Coller, M.~L. Loh, J.~R. Downing, M.~A. Caligiuri, et~al., Molecular
  classification of cancer: class discovery and class prediction by gene
  expression monitoring, science 286~(5439) (1999) 531--537.

\bibitem{squires2017non}
S.~Squires, L.~Montesdeoca, A.~Pr{\"u}gel-Bennett, M.~Niranjan, Non-negative
  matrix factorization with exogenous inputs for modeling financial data, in:
  International Conference on Neural Information Processing, Springer, 2017,
  pp. 873--881.

\end{thebibliography}

\end{document}